\def\paperlanguage{}
\pdfoutput=1 

\documentclass[letterpaper, 10 pt, conference]{ieeeconf}  

\usepackage{bm}
\usepackage{cite}
\include{preamble}

\newcommand{\ctext}[1]{\raise0.2ex\hbox{\textcircled{\scriptsize{#1}}}}

\IEEEoverridecommandlockouts                              

\title{\LARGE \textbf
  {
    \switchlanguage%
    {%
      PIMBS: Efficient Body Schema Learning for Musculoskeletal Humanoids with Physics-Informed Neural Networks
    }%
    {%
      PIMBS: PINNsによる筋骨格ヒューマノイドの効率的な身体図式学習
    }%
  }
}

\author{Kento Kawaharazuka$^{1}$, Takahiro Hattori$^{1}$, Keita Yoneda$^{1}$, and Kei Okada$^{1}$
  \thanks{$^{1}$ The authors are with the Department of Mechano-Informatics, Graduate School of Information Science and Technology, The University of Tokyo, 7-3-1 Hongo, Bunkyo-ku, Tokyo, 113-8656, Japan.
    {\texttt\small [kawaharazuka, t-hattori, yoneda, k-okada]@jsk.imi.i.u-tokyo.ac.jp}
  }
  \thanks{{\copyright} 2025 IEEE.  Personal use of this material is permitted.  Permission from IEEE must be obtained for all other uses, in any current or future media, including reprinting/republishing this material for advertising or promotional purposes, creating new collective works, for resale or redistribution to servers or lists, or reuse of any copyrighted component of this work in other works.
  }
}

\begin{document}

\maketitle
\thispagestyle{empty}
\pagestyle{empty}

\begin{abstract}
  \switchlanguage%
  {%
    Musculoskeletal humanoids are robots that closely mimic the human musculoskeletal system, offering various advantages such as variable stiffness control, redundancy, and flexibility.
    However, their body structure is complex, and muscle paths often significantly deviate from geometric models.
    To address this, numerous studies have been conducted to learn body schema, particularly the relationships among joint angles, muscle tension, and muscle length.
    These studies typically rely solely on data collected from the actual robot, but this data collection process is labor-intensive, and learning becomes difficult when the amount of data is limited.
    Therefore, in this study, we propose a method that applies the concept of Physics-Informed Neural Networks (PINNs) to the learning of body schema in musculoskeletal humanoids, enabling high-accuracy learning even with a small amount of data.
    By utilizing not only data obtained from the actual robot but also the physical laws governing the relationship between torque and muscle tension under the assumption of correct joint structure, more efficient learning becomes possible.
    We apply the proposed method to both simulation and an actual musculoskeletal humanoid and discuss its effectiveness and characteristics.
  }%
  {%
    筋骨格ヒューマノイドは人間の筋骨格系を詳細に模倣したロボットであり, 可変剛性制御や筋肉の冗長性, 柔軟性など多数の利点を有する.
    一方でその身体構造は複雑であり, 筋経路が幾何モデルと大きく異なる場合が多く, これまで様々な身体図式, 特に関節角度-筋張力-筋長の関係を学習する研究が行われてきた.
    これは一般的にロボット実機から得られたデータのみを用いて学習されるが, そのデータ収集には多大な労力が必要であり, かつデータ数が少ない場合には学習が困難である.
    そこで本研究では, Physics-Informed Neural Networkの考え方を筋骨格ヒューマノイドにおける身体図式学習に適用し, データ数が少ない場合でも高い精度で学習を行う手法を提案する.
    実機から得られたデータだけでなく, 関節構造を正しいと仮定した場合のトルクと筋張力の間の物理法則を活用することで, より効率的な学習が可能となる.
    提案手法をシミュレーションと筋骨格ヒューマノイド実機に適用し, その有効性と特性について考察する.
  }%
\end{abstract}

\section{INTRODUCTION}\label{sec:introduction}
\switchlanguage%
{%
  Various musculoskeletal humanoids that closely mimic human anatomy have been developed \cite{gravato2010ecce1, jantsch2013anthrob, asano2016kengoro, kawaharazuka2019musashi}.
  These robots offer several advantages, such as enabling variable stiffness control through nonlinear elastic elements \cite{jacobsen1990control, koganezawa1999stiffness} and allowing continued operation even when muscle rupture occurs due to muscle redundancy \cite{kawaharazuka2022redundancy}.
  Furthermore, due to their human-like structure, they have been developed as systems capable of directly implementing and verifying human reflex control and learning mechanisms \cite{endo1994reflex, marques2013reflex, liu2018stretch}.
  However, the complexity of their body structure often results in significant discrepancies between geometric models and the actual robot.
  The reasons for these discrepancies include variations in muscle paths and the inability of geometric models to accurately represent the muscles wrapped around the joints.

  To address this issue, various methods have been developed to learn the body schema of musculoskeletal structures, which captures the relationship between joint angle, muscle tension, and muscle length from actual data.
  In \cite{nakanishi2010estimation}, a method was proposed to construct a data table representing the relationship between joint angle and muscle length, which is updated using actual robot data.
  In \cite{ookubo2015learning}, polynomial approximation was used instead of a data table, and by differentiating it to obtain the muscle Jacobian, joint angle estimation was performed.
  In \cite{kawaharazuka2018online}, a neural network was used as an alternative to data tables and polynomial approximation, and it was applied to joint angle estimation and control.
  However, methods relying solely on joint angle and muscle length cannot account for factors such as wire elongation and nonlinear elastic elements.
  To address these limitations, other methods have been developed to learn the relationship between joint angle, muscle tension, and muscle length using actual robot data \cite{kawaharazuka2019longtime, kawaharazuka2020autoencoder}.
  On the other hand, obtaining a large amount of data from actual robots is challenging, making it desirable to develop methods that can accurately learn these relationships from a minimal amount of data.
}%
{%
  これまで人体構造を詳細に模倣した様々な筋骨格ヒューマノイドが開発されてきた\cite{gravato2010ecce1, jantsch2013anthrob, asano2016kengoro, kawaharazuka2019musashi}.
  これらのロボットは非線形弾性要素を活用した可変剛性制御が可能であったり\cite{jacobsen1990control, koganezawa1999stiffness}, 筋肉の冗長性によって筋破断時にも継続して動作が可能であったりと\cite{kawaharazuka2022redundancy}, 様々な利点を有している.
  また, 人間と同じ構造ゆえに人間の反射制御や学習機構を直接実装, 検証できるシステムとして長く開発されてきた\cite{endo1994reflex, marques2013reflex, liu2018stretch}.
  一方でその身体構造は複雑であり, 幾何モデルを構築しても, それは実際のロボットから大きくかけ離れている場合が多い.
  これは, 筋経路が一定に定まらなかったり, 関節に対する筋の巻き付きが幾何モデルで表現できなかったりと, その理由は様々である.

  この問題に対して, 筋骨格身体における身体図式と呼ばれる, 関節角度と筋張力, 筋長の関係性を実際のデータから学習する様々な手法が構築されてきた.
  \cite{nakanishi2010estimation}では関節角度と筋長の関係性をデータテーブルとして構築し, それを実機データから更新していくことで学習する手法を開発した.
  \cite{ookubo2015learning}ではデータテーブルの代わりに多項式近似を用い, これを微分して筋長ヤコビアンを求めることで関節角度推定を行っている.
  \cite{kawaharazuka2018online}ではデータテーブル, 多項式近似に代わるニューラルネットワークを用いて, これを関節角度推定や制御に活用した.
  しかし, 関節角度と筋長のみ情報ではワイヤや非線形弾性要素の伸びを考慮出来ておらず, これに対応するために, 関節角度-筋張力-筋長の関係性を学習するような手法も開発されている\cite{kawaharazuka2019longtime, kawaharazuka2020autoencoder}.
  一方で, 実際のロボットデータを大量に取得することは難しく, なるべく少数のデータから正確に関係性を学習できることが望ましい.
}%

\begin{figure}[t]
  \centering
  \includegraphics[width=0.9\columnwidth]{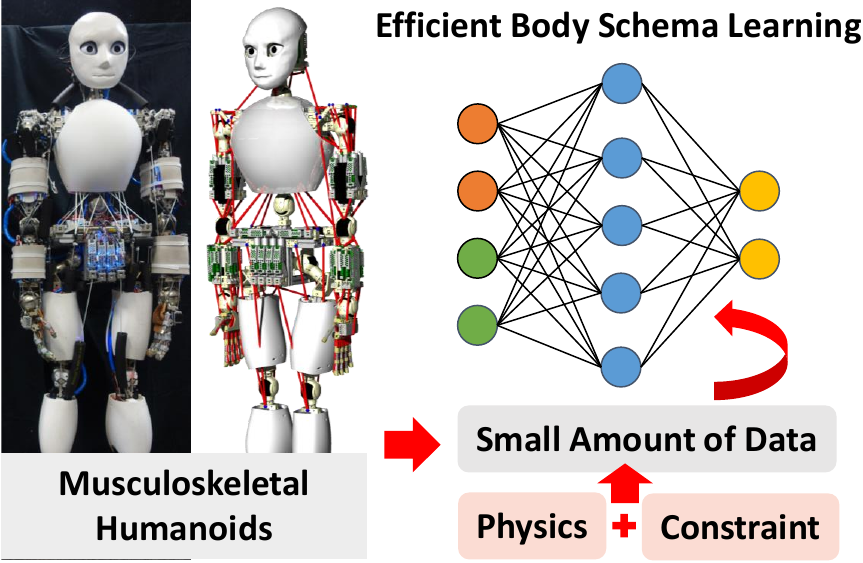}
  \vspace{-1.0ex}
  \caption{The concept of this study: in musculoskeletal humanoids, efficient body schema learning can be achieved by leveraging a small amount of data obtained from the actual robot along with the physical knowledge and constraints of the musculoskeletal system.}
  \label{figure:concept}
  \vspace{-4.0ex}
\end{figure}

\switchlanguage%
{
  In this study, we developed a method that enhances the neural network for the learning of the relationship between joint angle, muscle tension, and muscle length by incorporating constraints based on physical laws, enabling more efficient learning of the body schema (\figref{figure:concept}).
  We incorporate the concept of Physics-Informed Neural Networks (PINNs) \cite{raissi2019pinn}, which embeds the relationship between muscle tension and joint torque -- assuming that the joint geometric model is correct -- into the loss function.
  We refer to this approach as the Physics-Informed Musculoskeletal Body Schema, or PIMBS.
  By utilizing this method, not only the data points themselves but also their differential information can be obtained, allowing for more efficient body schema learning from fewer data points.
  We conducted experiments using both a musculoskeletal structure in simulation and an actual musculoskeletal humanoid to evaluate the proposed method and discuss its characteristics.

  The contributions of this study are summarized as follows.
  \begin{itemize}
    \item Proposal of PIMBS, a method for efficiently learning body schema from a small amount of data using physical laws
    \item Validation of the proposed method on a simulated musculoskeletal structure
    \item Application of the proposed method to an actual musculoskeletal humanoid and discussion of its characteristics
  \end{itemize}
  %
}%
{%
  そこで本研究では, これまでの関節角度-筋張力-筋長の関係性を学習するニューラルネットワークに物理的な法則による制約を加え, より効率的に身体図式を学習可能な手法を開発する(\figref{figure:concept}).
  Physics-Informed Neural Networks (PINNs) \cite{raissi2019pinn}の考え方を導入し, 関節に関する幾何モデルが正しいと仮定した際の筋張力と関節トルクの関係性も損失関数として埋め込む.
  我々はこれを, Physics-Informed Musculoskeletal Body Schema, PIMBSと呼んでいる.
  これにより, 単なるデータ点のみの情報だけでなく, それらの微分の情報が得られるため, より少ないデータ点から効率的に身体図式を学習することができる.
  提案手法についてシミュレーション上における筋骨格構造と筋骨格ヒューマノイド実機を用いて実験を行い, その特性について議論する.
  本研究の貢献について以下に列挙する.
  \begin{itemize}
    \item 少量のデータから物理法則を用いて身体図式を効率的に学習するPIMBSの提案
    \item シミュレーション上の筋骨格構造における提案手法の有効性検証
    \item 実機筋骨格ヒューマノイドにおける提案手法の適用とその特性の考察
  \end{itemize}

  本研究の構成は以下である.
  \secref{sec:proposed}では, 一般的な筋骨格構造と身体図式学習の基礎について説明し, 提案する物理法則の導入について述べる.
  \secref{sec:experiment}では, 2自由度4筋の筋骨格シミュレーションと実際の筋骨格ヒューマノイドの左腕を用いて実験を行い, 基礎的な身体図式学習と提案手法の性能を比較する.
  \secref{sec:discussion}で実験結果について議論し, 最後に\secref{sec:conclusion}で結論を述べる.
}%

\begin{figure}[t]
  \centering
  \includegraphics[width=0.9\columnwidth]{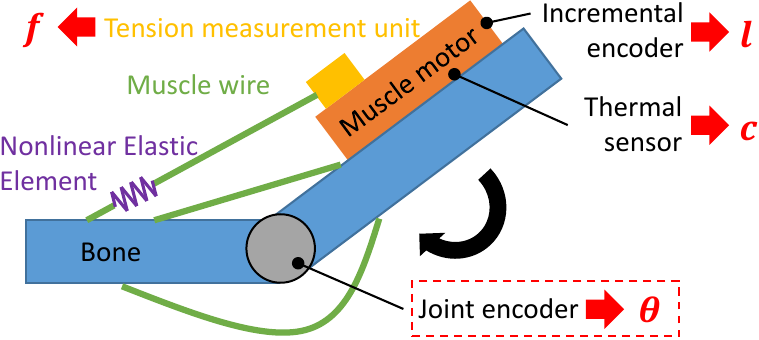}
  \vspace{-1.0ex}
  \caption{The basic structure of musculoskeletal humanoids: muscle motors, equipped with incremental encoders, thermal sensors, and tension measurement units, are attached to the skeleton. Muscle wires extend from the muscle motors and are connected to the skeleton via nonlinear elastic elements. The presence of joint encoders depends on the specific robot.}
  \label{figure:basic-structure}
  \vspace{-3.0ex}
\end{figure}

\section{PIMBS: Efficient Body Schema Learning for Musculoskeletal Humanoids} \label{sec:proposed}

\subsection{Basic Structure of Musculoskeletal Humanoids} \label{subsec:basic}
\switchlanguage%
{%

  The general musculoskeletal structure is explained below.
  As shown in \figref{figure:basic-structure}, the musculoskeletal structure typically consists of a skeleton, joints, and wires that simulate muscles.
  The wires are wound around pulleys attached to the motor, pass through the muscle tension measurement unit, and are connected to the skeleton.
  Nonlinear elastic elements are attached to the ends of the wires, enabling variable stiffness control.
  Muscle length $l$ is obtained from the encoder attached to the motor, and muscle tension $f$ is measured using a loadcell installed in the muscle tension measurement unit.
  Joint angle $\theta$ may or may not be directly measured depending on the actual robot.
  However, even when joint angle measurements are unavailable, they can be estimated with reasonable accuracy by combining muscle length variations with data from visual sensors \cite{kawaharazuka2018online}.

  When the wires do not stretch and nonlinear elasticity is absent, the following relationship holds using the function $\bm{h}$:
  \begin{align}
    \bm{l} = \bm{h}(\bm{\theta}) \label{eq:bodyschema-1}
  \end{align}
  The derivative of this function with respect to $\bm{\theta}$ is called the muscle Jacobian and is denoted by $\bm{G}$:
  \begin{align}
    \bm{G}(\bm{\theta}) = \frac{\partial\bm{h}(\bm{\theta})}{\partial\bm{\theta}}
  \end{align}

  Next, consider the case where the wires can stretch and nonlinear elastic elements are installed.
  Let the amount that the wire is stretched, which varies with joint angle and muscle tension, be denoted by $\Delta\bm{n}(\bm{\theta}, \bm{f})$.
  The following relationship holds:
  \begin{align}
    \bm{l}^{geom} &= \bm{h}^{geom}(\bm{\theta}) \\
    \bm{l} = \bm{h}(\bm{\theta}, \bm{f}) &= \bm{h}^{geom}(\bm{\theta}) + \Delta\bm{n}(\bm{\theta}, \bm{f}) \label{eq:bodyschema-2}
  \end{align}
  where $\bm{l}^{geom}$ represents the geometric wire length corresponding to the joint angle, and the actual muscle length $\bm{l}$ obtained from the motor is further adjusted by the amount $\Delta\bm{n}$.
  Note that the joint angle is included in $\Delta\bm{n}(\bm{\theta}, \bm{f})$ because, in general, the amount that the wire is stretched depends on the total wire length $\bm{l}^{geom}$.
  The muscle Jacobian is defined as the derivative of $\bm{l}^{geom}$ with respect to $\bm{\theta}$.
  However, since the variation of $\Delta\bm{n}$ with respect to $\bm{\theta}$ is generally much smaller than the variation of $\bm{h}^{geom}(\bm{\theta})$, the following approximation is often used:
  \begin{align}
    \bm{G}(\bm{\theta}) = \frac{\partial\bm{h}^{geom}(\bm{\theta})}{\partial\bm{\theta}} \simeq \frac{\partial\bm{h}(\bm{\theta}, \bm{f})}{\partial\bm{\theta}}
  \end{align}

}%
{%
  一般的な筋骨格構造について説明する.
  筋骨格構造は\figref{figure:basic-structure}に示すように, 基本的に骨格と関節, 筋肉を模したワイヤから構成される.
  ワイヤはモータの先端に着いたプーリに巻かれ, 筋張力測定ユニットを通って骨格に接続する.
  ワイヤの先端には非線形弾性が取り付けられており, これが可変剛性制御を可能にする.
  モータに取り付けられたエンコーダから筋長$l$が, 筋張力測定ユニットに取り付けられた圧力センサから筋張力$f$が得られる.
  関節角度$\theta$はロボットによっては, 得られること, 得られないことがあるが, 得られない場合でも筋長変化と視覚センサを併用することで, ある程度正確に推定することが可能である\cite{kawaharazuka2018online}.

  ワイヤが伸びなく, 非線形弾性を持たない場合は関数$\bm{h}$を用いて以下の関係式が成り立つ.
  \begin{align}
    \bm{l} = \bm{h}(\bm{\theta}) \label{eq:bodyschema-1}
  \end{align}
  これを$\bm{\theta}$で微分したものを筋長ヤコビアンと呼び$\bm{G}$で表す.
  \begin{align}
    \bm{G}(\bm{\theta}) = \frac{\partial\bm{h}(\bm{\theta})}{\partial\bm{\theta}}
  \end{align}

  ワイヤが伸びたり, 非線形弾性要素が取り付けられた場合について考える.
  関節角度と筋張力によって変化するそれらの伸びを$\Delta\bm{n}(\bm{\theta}, \bm{f})$とすると, 以下の関係が成り立つ.
  なお, 関節角度が入るのは, 一般的にワイヤの伸びが現在のワイヤ長さに依存するためである.
  \begin{align}
    \bm{l}^{geom} &= \bm{h}^{geom}(\bm{\theta}) \\
    \bm{l} = \bm{h}(\bm{\theta}, \bm{f}) &= \bm{h}^{geom}(\bm{\theta}) + \Delta\bm{n}(\bm{\theta}, \bm{f}) \label{eq:bodyschema-2}
  \end{align}
  ここで, $\bm{l}^{geom}$は関節角度に対応した幾何的なワイヤの長さを表し, 実際にモータから得られる$\bm{l}$はそこからさらに$\Delta\bm{n}$だけ巻き取った値となる.
  筋長ヤコビアンはこの$\bm{l}^{geom}$を$\bm{\theta}$で微分したものであるが, $\bm{\theta}$に応じた$\Delta\bm{n}$の変化は$\bm{h}^{geom}(\bm{\theta})$の変化に比べて十分小さいと考えて, 以下のように近似する場合が多い.
  \begin{align}
    \bm{G}(\bm{\theta}) = \frac{\partial\bm{h}^{geom}(\bm{\theta})}{\partial\bm{\theta}} \simeq \frac{\partial\bm{h}(\bm{\theta}, \bm{f})}{\partial\bm{\theta}}
  \end{align}
}%

\begin{figure}[t]
  \centering
  \includegraphics[width=0.9\columnwidth]{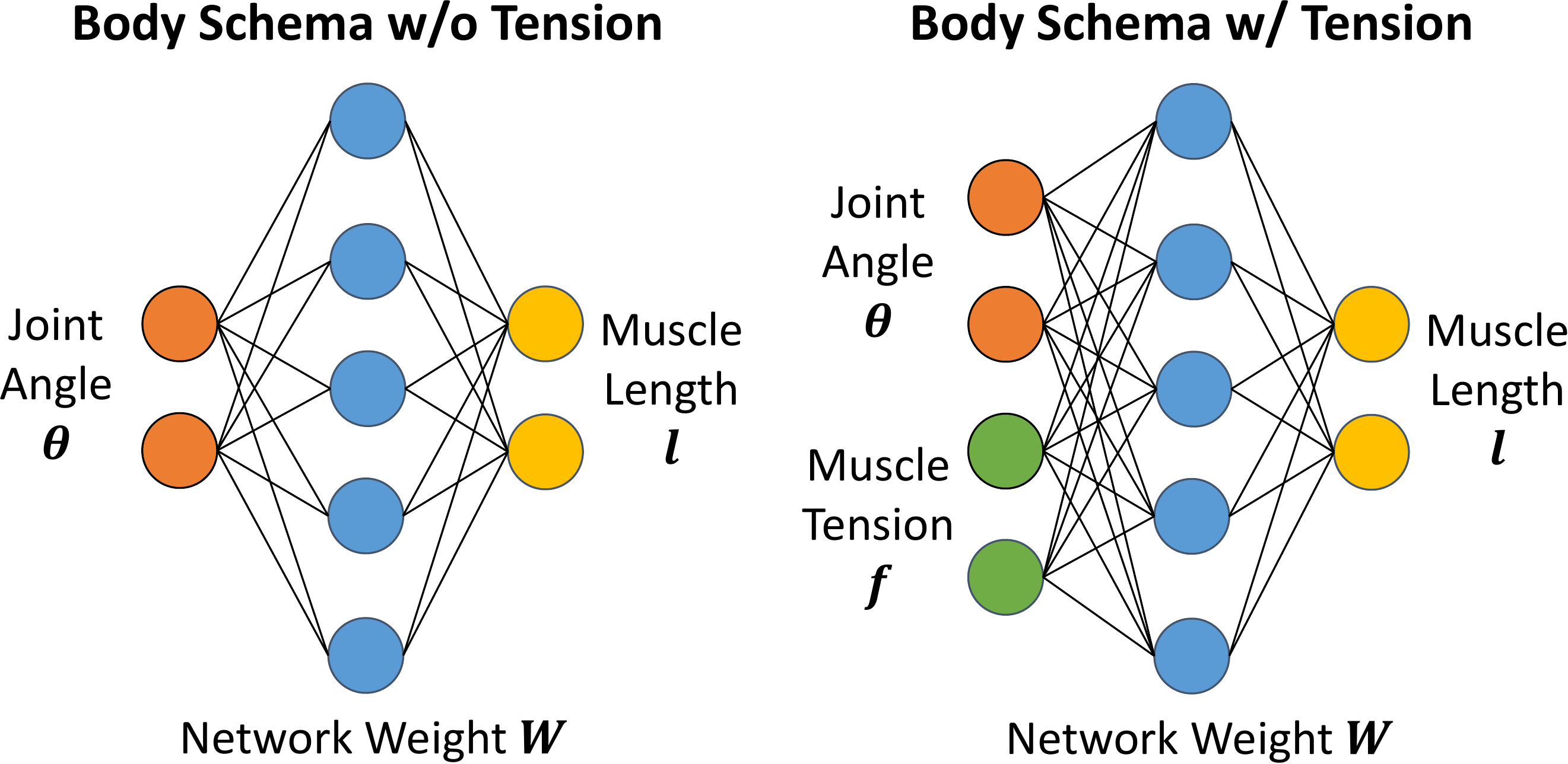}
  \vspace{-1.0ex}
  \caption{The neural network structures for basic body schema learning: the input consists of joint angles, and in some cases, muscle tension; the output is the muscle length.}
  \label{figure:basic-learning}
  \vspace{-3.0ex}
\end{figure}

\subsection{Basic Body Schema Learning for Musculoskeletal Humanoids} \label{subsec:learning}
\switchlanguage%
{%
  The general method for learning the body schema of a musculoskeletal humanoid is described below.
  Basically, the body schema consists of a network that outputs muscle length from joint angle or from joint angle and muscle tension, as shown in \figref{figure:basic-learning}.
  In other words, the network learns $\bm{h}$ in \equref{eq:bodyschema-1} or \equref{eq:bodyschema-2} as a neural network.

  The learning process is simple.
  The actual robot is moved, and the data at each moment, $D_{i} = (\bm{\theta}^{data}{i}, \bm{f}^{data}{i}, \bm{l}^{data}_{i})$, is collected.
  The robot can be moved either by applying random muscle tension or by commanding muscle length based on a simple geometric model and adding muscle tension control to eliminate muscle slack and antagonistic effects.
  It is assumed that there is no muscle slack.
  It should be noted that the geometric model used here often differs significantly from the actual robot.
  The collected data for learning is denoted as $D^{train}$, with the total number of data points represented by $N^{train}$.
  Using these data, learning is performed with the mean squared error as the loss function, as shown below:
  \begin{align}
    \bm{l}^{pred} &= \bm{h}(\bm{\theta}^{data}) \;\;\; \textrm{OR} \;\;\; \bm{h}(\bm{\theta}^{data}, \bm{f}^{data}) \\
    L_{basic} &= \frac{1}{N^{train}}\sum(\bm{l}^{pred}-\bm{l}^{data})^{2} \label{eq:l-basic}
  \end{align}
  By using the trained $\bm{h}$, the robot can obtain the desired muscle length $\bm{l}$ by inputting either $\bm{\theta}$ or $(\bm{\theta}, \bm{f})$.
  Furthermore, muscle Jacobian $\bm{G}$, derived by differentiating $\bm{h}$, can be used for state estimation and torque control.
  The muscle length $\bm{l}$ is expressed as a relative change, with its value at $\bm{\theta} = \bm{0}$ set to $\bm{0}$.

  This general body schema learning is highly effective, allowing for increasingly accurate learning as more data is collected.
  However, in practice, it is often difficult to obtain large amounts of data, making it desirable to learn with fewer data points.
  To address this, we incorporate the concept of Physics-Informed Neural Networks (PINNs), which leverage differential equations expressed through network derivatives.
  We refer to this network as the Physics-Informed Musculoskeletal Body Schema (PIMBS).
}%
{%
  筋骨格ヒューマノイドにおける一般的な身体図式学習の方法について述べる.
  基本的に, 身体図式は\figref{figure:basic-learning}に示すような関節構造から筋長, または関節角度と筋張力から筋長を出力するネットワークによって構成される.
  つまり, \equref{eq:bodyschema-1}または\equref{eq:bodyschema-2}における$\bm{h}$をニューラルネットワークとして学習する.

  学習の方法はいたって単純である.
  ロボットを動かし, そのときのデータ$D_{i} = (\bm{\theta}^{data}_{i}, \bm{f}^{data}_{i}, \bm{l}^{data}_{i})$を取得していく.
  これは, ランダムな張力を加えながら動かしても良いし, 簡易的に構築した幾何モデルから筋長を指令し, 筋の拮抗や緩みを無くすような張力制御を追加しながら動かしても良い.
  なお, 筋の緩みはないと仮定する.
  そして, ここでいう幾何モデルは実際のロボットとは大きく異なる場合が多い.
  ここで得られた学習用のデータを$D^{train}$とし, そのデータ数を$N^{train}$とする.
  それらのデータを用いて以下のようにmean squared errorを誤差として学習を行う.
  \begin{align}
    \bm{l}^{pred} &= \bm{h}(\bm{\theta}^{data}) \;\;\; \textrm{OR} \;\;\; \bm{h}(\bm{\theta}^{data}, \bm{f}^{data}) \\
    L_{basic} &= \frac{1}{N^{train}}\sum(\bm{l}^{pred}-\bm{l}^{data})^{2} \label{eq:l-basic}
  \end{align}
  ここで得られた$\bm{h}$によって, $\bm{\theta}$または$(\bm{\theta}, \bm{f})$を入力すればロボットが実現すべき筋長$\bm{l}$が得られるし, これを微分することで得られた筋長ヤコビアン$\bm{G}$を用いて張力制御を行うこともできる.
  なお, 筋長$\bm{l}$は$\bm{\theta}=\bm{0}$における値を$\bm{0}$として, そこからの相対変化として表現される.

  一般的な身体図式学習は非常に効果的でデータを集めれば集めるほど精度の高い学習が可能である.
  一方で, 一般に大量のデータを取得することは難しく, より少ないデータ数で学習できることが望ましい.
  ここで我々は, ネットワークの微分で表現した微分方程式を活用するPhysics-Informed Neural Networks (PINNs)の考え方を導入する.
  このネットワークをPhysics-Informed Musculoskeletal Body Schema, PIMBSと呼ぶ.
}%

\begin{figure}[t]
  \centering
  \includegraphics[width=0.85\columnwidth]{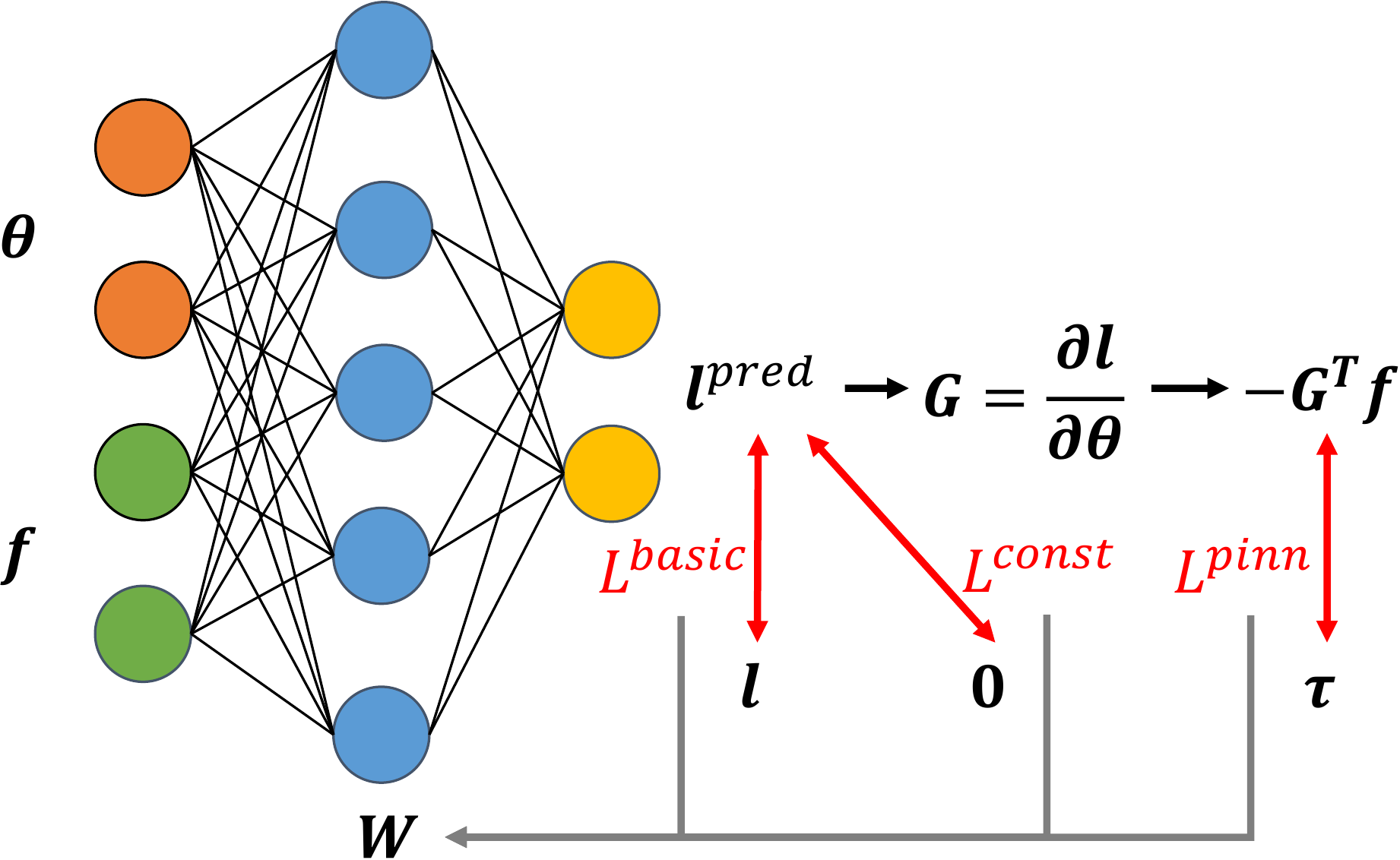}
  \vspace{-1.0ex}
  \caption{The training scheme of Physics-Informed Musculoskeletal Body Schema, PIMBS.}
  \label{figure:pinn}
  \vspace{-3.0ex}
\end{figure}

\subsection{Physics-Informed Body Schema Learning for Musculoskeletal Humanoids} \label{subsec:pinn}
\switchlanguage%
{%
  An overview of the training scheme of PIMBS is shown in \figref{figure:pinn}.
  In body schema learning, it is generally assumed that the joint structure (i.e., the positional relationships between joints and links, as well as link weights) is correct, and only the relationship between joints and muscles is learned.
  Assuming the joint structure is correct, the gravity compensation torque $\bm{\tau}$ corresponding to the current joint angle $\bm{\theta}$ can be calculated.
  In musculoskeletal structures, the following relationship generally holds:
  \begin{align}
    \bm{\tau} = -\bm{G}^{T}(\bm{\theta})\bm{f}
  \end{align}
  In other words, by utilizing $\bm{G}$, which is the derivative of the function $\bm{h}$ represented by the neural network, it is possible to introduce an additional loss function for learning.
  Therefore, the following loss can be defined:
  \begin{align}
    \bm{G}^{pred} &= \left. \frac{\partial\bm{h}(\bm{\theta})}{\partial\bm{\theta}} \right|_{\bm{\theta}=\bm{\theta}^{data}} \;\;\; \textrm{OR} \;\;\; \left. \frac{\partial\bm{h}(\bm{\theta}, \bm{f})}{\partial\bm{\theta}} \right|_{\bm{\theta}=\bm{\theta}^{data}, \bm{f}=\bm{f}^{data}} \\
    L_{pinn} &= \frac{1}{N^{train}}\sum(\bm{G}^{pred, T}\bm{f}^{data}+\bm{\tau}^{data})^{2} \label{eq:l-pinn}
  \end{align}
  where $\bm{\tau}^{data}$ represents the gravity compensation torque for the joint angle $\bm{\theta}^{data}$.
  Additionally, constraint conditions can also be described.
  When $\bm{\theta}=\bm{0}$ and $\bm{f}=\bm{0}$, $\bm{l}$ should be $\bm{0}$, allowing for the definition of the following loss:
  \begin{align}
    \bm{l}^{pred} &= \bm{h}(\bm{0}) \;\;\; \textrm{OR} \;\;\; \bm{h}(\bm{0}, \bm{0}) \\
    L_{const} &= (\bm{l}^{pred}-\bm{0})^2 \label{eq:l-const}
  \end{align}
  By additionally utilizing these losses $L_{pinn}$ and $L_{const}$ alongside $L_{basic}$, the body schema can be learned more efficiently.
  For the experiments, we compare the learning performance by varying the loss function $L_{train}$ among $L_{basic}$ (\textbf{Basic}), $L_{basic}+L_{const}$ (\textbf{Basic+Const}), $L_{basic}+\alpha L_{pinn}$ (\textbf{Basic+PINN}), and $L_{basic}+L_{const}+\alpha L_{pinn}$ (\textbf{Basic+Const+PINN}).
  Here, $\alpha$ is the weight for the physical laws.
  Among these loss functions, \textbf{Basic} is the conventional approach that has been used in almost all previous studies.
  Although there are other methods such as polynomial approximation and data table-based approaches, they are not included as baselines in this study due to the difficulty of making a fair comparison with neural networks.
  Note that all displayed loss values are scaled by a factor of $1.0 \times 10^{5}$ for clearer visualization.
}%
{%
  \figref{figure:pinn}にPIMBSの学習の概要を示す.
  身体図式の学習の際には, 一般に関節構造(関節とリンクの位置関係やリンクの重さ)は正しいとして, 関節と筋の間の関係性のみを学習している.
  この関節構造を正しいとすると, 現在の関節角度$\bm{\theta}$に対応した重力補償トルク$\bm{\tau}$を計算することができる.
  そして, 筋骨格構造には一般に以下の関係が成り立つ.
  \begin{align}
    \bm{\tau} = -\bm{G}^{T}(\bm{\theta})\bm{f}
  \end{align}
  つまるところ, これまで身体図式は\equref{eq:l-basic}のみを用いて学習されてきたが, 実はニューラルネットワークで表された関数$\bm{h}$の微分である$\bm{G}$を用いることで, もう一つ別の損失関数を用いて学習することも可能なのである.
  よって, 以下のような損失を定義することができる.
  \begin{align}
    \bm{G}^{pred} &= \left. \frac{\partial\bm{h}(\bm{\theta})}{\partial\bm{\theta}} \right|_{\bm{\theta}=\bm{\theta}^{data}} \;\;\; \textrm{OR} \;\;\; \left. \frac{\partial\bm{h}(\bm{\theta}, \bm{f})}{\partial\bm{\theta}} \right|_{\bm{\theta}=\bm{\theta}^{data}, \bm{f}=\bm{f}^{data}} \\
    L_{pinn} &= \frac{1}{N^{train}}\sum(\bm{G}^{pred, T}\bm{f}^{data}+\bm{\tau}^{data})^{2} \label{eq:l-pinn}
  \end{align}
  ここで, $\bm{\tau}^{data}$は$\bm{\theta}^{data}$における重力補償トルクを表す.
  加えて, 拘束条件を記述することもできる.
  $\bm{\theta}=\bm{0}$かつ$\bm{f}=\bm{0}$のとき, $\bm{l}=\bm{0}$であるべきなため, 以下の損失を定義することができる.
  \begin{align}
    \bm{l}^{pred} &= \bm{h}(\bm{0}) \;\;\; \textrm{OR} \;\;\; \bm{h}(\bm{0}, \bm{0}) \\
    L_{const} &= (\bm{l}^{pred}-\bm{0})^2 \label{eq:l-const}
  \end{align}
  これら$L_{pinn}, L_{const}$を$L_{basic}$に追加で活用することで, より効率的に身体図式を学習することができる.
  学習する際の損失関数$L_{train}$を$L_{basic}$ (\textbf{Basic}), $L_{basic}+L_{const}$ (\textbf{Basic+Const}), $L_{basic}+\alpha L_{pinn}$ (\textbf{Basic+PINN}), $L_{basic}+L_{const}+\alpha L_{pinn}$ (\textbf{Basic+Const+PINN})と変化させながら比較実験を行う.
  ここで, $\alpha$は物理法則に対する重みである.
  これらの損失関数のうち, \textbf{Basic}がほとんど全ての研究で用いられている既存の方式である.
  他にも多項式近似やデータテーブルを用いた方法があるが, ニューラルネットワークとのフェアな比較は難しいため, 今回は比較対象としていない.
  なお, 表示する損失の値は全て$1.0 \times 10^{5}$をかけて分かりやすい値に変換して表示している.
}%

\begin{figure}[t]
  \centering
  \includegraphics[width=0.9\columnwidth]{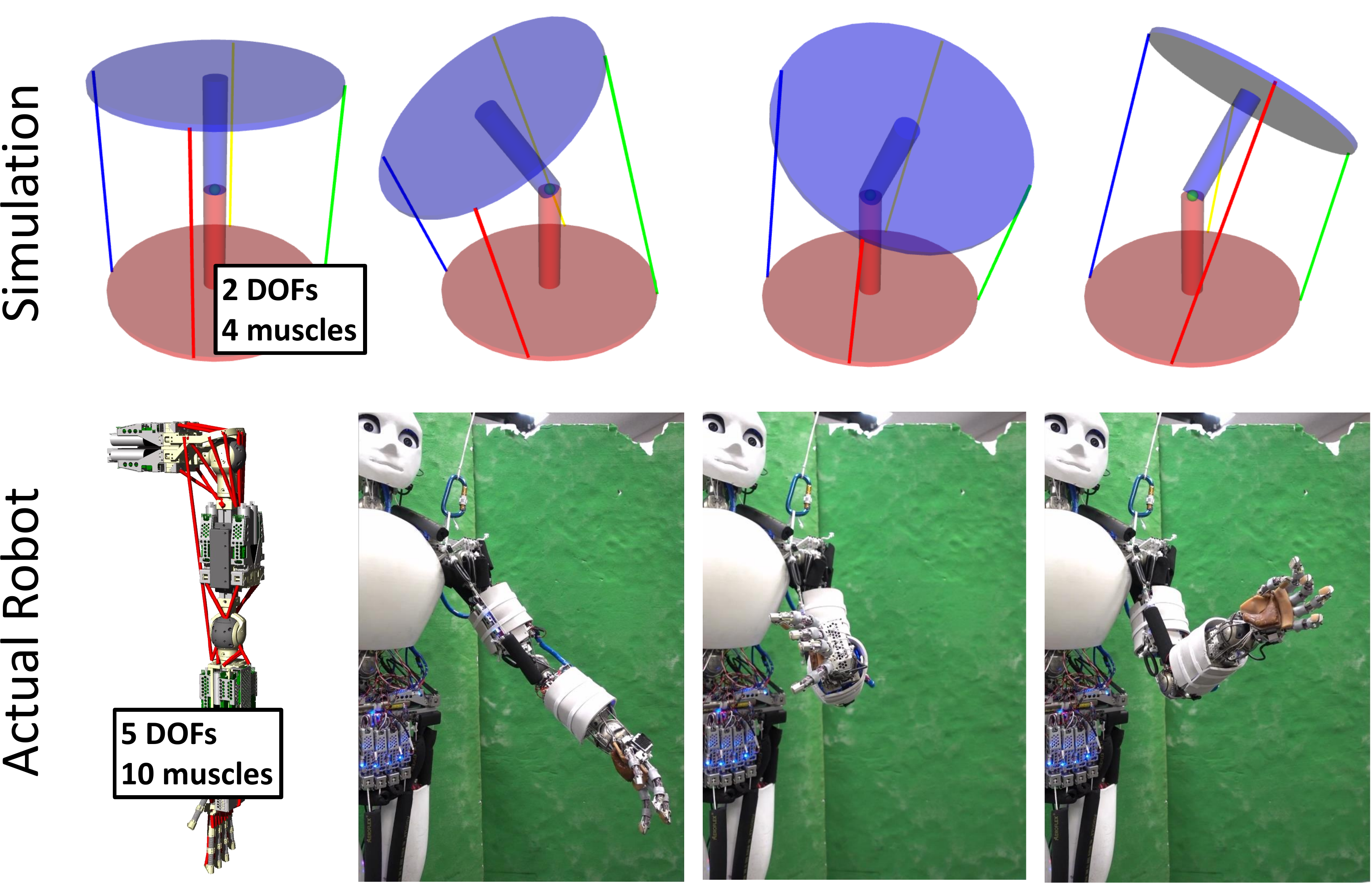}
  \vspace{-1.0ex}
  \caption{The musculoskeletal structures handled in our experiments: 2-DOF 4-muscle musculoskeletal simulation and 5-DOF 10-muscle left arm of the musculoskeletal humanoid.}
  \label{figure:exp-setup}
  \vspace{-3.0ex}
\end{figure}

\section{Experiments} \label{sec:experiment}

\subsection{Experimental Setup} \label{subsec:exp-setup}
\switchlanguage%
{%
  The experimental setup of this study is described here.
  In this study, we use a 2-DOF 4-muscle musculoskeletal simulation, and the 5-DOF 10-muscle left arm of a musculoskeletal humanoid, as shown in \figref{figure:exp-setup}.

  First, for the 2-DOF 4-muscle simulation, experiments are conducted using two formulations defined by \equref{eq:bodyschema-1} and \equref{eq:bodyschema-2}.
  To simplify understanding, the formulation in \equref{eq:bodyschema-1} is referred to as the AL-Map (angle-length map), and the one in \equref{eq:bodyschema-2} is referred to as the ATL-Map (angle-tension-length map).
  Initially, training data for $(\bm{\theta}, \bm{f}, \bm{l}, \bm{\tau})$ is generated in simulation.
  The joint angle is randomly varied within the range of $[-0.5, 0.5]$ [rad].
  Once $\bm{\theta}$ is determined, the geometric muscle length $\bm{l}^{geom}$ and the gravity compensation torque $\bm{\tau}$ are calculated geometrically.
  During this process, the stretch $\Delta{n}$ of each nonlinear elastic element is obtained using the following relationship, and $\bm{l}$ is then calculated in the same form as \equref{eq:bodyschema-2}:
  \begin{align}
    f=e^{K\Delta{n}}-1
  \end{align}
  where $K$ is a coefficient, set to 1000 in this study.
  The muscle tension $\bm{f}$ is generated by solving the following quadratic programming problem:
  \begin{align}
    \underset{\bm{f}}{\textrm{minimize}}&\;\;\;\;\;\;\;\;\;\;\;\;\;\;\;\;\;\bm{f}^{T}\bm{W}\bm{f}\label{eq:tension-calc}\\
    \textrm{subject to}&\;\;\;\;\;\;\;\;\;\;\;\; \bm{\tau} = -\bm{G}^{T}(\bm{\theta})\bm{f}\nonumber\\
    &\;\;\;\;\;\;\;\;\;\;\;\; \bm{f} \geq \bm{f}^{min} \nonumber
  \end{align}
  where $W$ represents the weighting matrix, which is set to the identity matrix in this study.
  Additionally, $f^{min}$ represents the minimum muscle tension.
  In experiments related to the AL-Map, $f^{min}$ is set to a fixed value of $0$ [N].
  This allows $\bm{\tau}$, $\bm{f}$, and $\bm{l}$ to be uniquely determined from $\bm{\theta}$, enabling the construction of a body schema in the form of the AL-Map.
  The formulation in \equref{eq:bodyschema-1} assumes inextensible wires, but a similar formulation can be applied even when wires are stretched, as long as $\bm{l}$ can be uniquely determined from $\bm{\theta}$.
  On the other hand, in experiments related to the ATL-Map, variations in $\bm{f}$ are also accepted, allowing $f^{min}$ to be randomly varied within the range of [10, 300] N during the calculation of $\bm{f}$.
  In this way, experiments are conducted on a more complex body schema that takes both $\bm{\theta}$ and $\bm{f}$ as inputs.

  The following experiments were conducted using the left arm of a musculoskeletal humanoid \cite{kawaharazuka2019musashi}, which has 5 joints and 10 muscles.
  The joint angle of the left arm was controlled and held stationary repeatedly in a random manner based on a geometric model.
  During stationary phases, muscle tension control was applied to eliminate muscle slack, and data was collected in this process.
  Since each muscle is equipped with a nonlinear elastic element, the body schema was learned in the form of the ATL-Map.

  Details of the body schema learning are as follows.
  The body schema is constructed using a neural network with three fully connected layers.
  The input dimension is $N$ or $N+M$, the number of hidden units is 1000, and the output dimension is $M$, where $N$ represents the number of joints and $M$ represents the number of muscles.
  A hyperbolic tangent function is applied after the first and second fully connected layers.
  The batch size is set to match $N^{train}$, the number of epochs is 20000, and Adam \cite{kingma2015adam} is used as the update rule.
  The weight $\alpha$ for the physical laws is fixed at $1.0 \times 10^{-5}$ in simulation, while in the actual robot experiments, it is varied among $1.0 \times 10^{\{-5, -6, -7, -8\}}$ (details will be discussed in \secref{subsec:act-exp}).
  Experiments were performed by varying the number of training data $N^{train}$ and the loss function, between \textbf{Basic}, \textbf{Basic+Const}, \textbf{Basic+PINN}, and \textbf{Basic+Const+PINN}.
  In simulations, a separate evaluation dataset $D^{eval}$ consisting of 1000 randomly generated points was prepared.
  The evaluation value $L_{eval}$ is defined as the result of calculating \equref{eq:l-basic} using $D^{eval}$.
  In the actual robot experiments, out of 496 data points obtained from the actual robot, the data excluding $N^{train}$ points was used as $D^{eval}$, and $L_{eval}$ was calculated.
  The final model selected is the one with the minimum $L_{basic}$ obtained from the training data during all epochs, and the corresponding evaluation value is denoted as $L^{best}_{eval}$.
  In this study, the performance of different learning methods is compared based on the change in $L_{eval}$ during the learning process and the final value of $L^{best}_{eval}$.

}%
{%
  本研究の実験設定について述べる.
  本研究では\figref{figure:exp-setup}に述べる2関節4筋の筋骨格シミュレーションと, 5関節10筋の筋骨格ヒューマノイドの左腕を扱う.

  まず2関節4筋のロボットにおいて, \equref{eq:bodyschema-1}と\equref{eq:bodyschema-2}の2つの構成を扱って実験を進める.
  分かりやすいように, \equref{eq:bodyschema-1}の定式化をAL-Map (angle-length map), \equref{eq:bodyschema-2}の定式化をATL-Map (angle-tension-length map)と呼ぶ.
  まず, シミュレーション上で$(\bm{\theta}, \bm{f}, \bm{l}, \bm{\tau})$の訓練データを生成する.
  各関節の角度を$[-0.5, 0.5]$ [rad]の範囲でランダムに変化させる.
  $\bm{\theta}$が決まれば幾何的に$\bm{l}^{geom}$と重力補償トルク$\bm{\tau}$が求まる.
  この際, 各非線形弾性要素の伸び$\Delta{n}$を以下の関係式から得た上で, \equref{eq:bodyschema-2}と同じ形で$\bm{l}$を求めている.
  \begin{align}
    f=e^{K\Delta{n}}-1
  \end{align}
  ここで$K$は係数であり, 本研究では1000とした.
  $\bm{f}$については以下の二次計画法を解くことで生成する.
  \begin{align}
    \underset{\bm{f}}{\textrm{minimize}}&\;\;\;\;\;\;\;\;\;\;\;\;\;\;\;\;\;\bm{f}^{T}\bm{W}\bm{f}\label{eq:tension-calc}\\
    \textrm{subject to}&\;\;\;\;\;\;\;\;\;\;\;\; \bm{\tau} = -\bm{G}^{T}(\bm{\theta})\bm{f}\nonumber\\
    &\;\;\;\;\;\;\;\;\;\;\;\; \bm{f} \geq \bm{f}^{min} \nonumber
  \end{align}
  ここで$W$は重み行列を表し, 本研究では単位行列とする.
  また, $f^{min}$は筋張力の最小値であり, AL-Mapに関する実験では, $f^{min}=0$ [N]と固定値に設定している.
  これによって, $\bm{\theta}$から$\bm{\tau}$, $\bm{f}$, $\bm{l}$が一意に求まるため, AL-Mapの形で身体図式を構築することができる.
  \equref{eq:bodyschema-1}は伸びのないワイヤを仮定して定義したが, 伸びがある場合でも$\bm{\theta}$から$\bm{l}$が一意に定まれば同様の定式化が可能である.
  一方で, ATL-Mapに関する実験では, $\bm{f}$に関する変化も受け付けることができるため, $f^{min}$を[10, 300] Nの範囲でランダムに変化させながら$\bm{f}$を計算している.
  つまり, $\bm{\theta}$と$\bm{f}$を入力とするより複雑な身体図式に関する実験を行う.

  次に筋骨格ヒューマノイド \cite{kawaharazuka2019musashi}の5関節10筋の左腕を用いて実験を行う.
  この左腕の関節角度を, 幾何モデルに基づいてランダムに制御と静止を繰り返し, 静止時に筋の緩みを無くすような筋張力制御を行いながらデータを収集する.
  各筋肉には非線形弾性要素が取り付けられているため, ATL-Mapの形で身体図式を学習する.

  身体図式学習の詳細について述べる.
  身体図式は全て3層の全結合層からなるニューラルネットワークによって構成されており, 入力次元は$N$または$N+M$, 隠れ層のユニット数は1000, 出力次元は$M$となっている ($N$は関節数, $M$は筋数を表す).
  1層目と2層目の全結合層の後にはhyperbolic tangentが適用される.
  バッチ数は$N^{train}$と一致させ, エポック数は20000, update ruleはAdam \cite{kingma2015adam}を用いる.
  物理法則への重みである$\alpha$は, シミュレーションでは$1.0 \times 10^{-5}$で統一し, 実機実験では$1.0 \times 10^{\{-5, -6, -7, -8\}}$とした(詳細については後に述べる).
  データ数$N^{train}$, そして損失関数を\textbf{Basic}, \textbf{Basic+Const}, \textbf{Basic+PINN}, \textbf{Basic+Const+PINN}に変化させながら実験を行う.
  この際, シミュレーションにおいては, 別に1000点のデータをランダムに生成した評価用データセット$D^{eval}$を用意し, これを用いて\equref{eq:l-basic}を計算した際の値を評価値$L_{eval}$とする.
  実機実験においては, 実機から得られた496点のデータのうち, $N^{train}$個のデータを引いたものを$D^{eval}$とし, $L_{eval}$を計算する.
  最終的に選ぶモデルは, 全エポック内で訓練データに対して最小の$L_{basic}$が得られた際のモデルとし, このときの評価値を$L^{best}_{eval}$とする.
  本研究では, 基本的に学習過程における$L_{eval}$の変化と, 最終的な$L^{best}_{eval}$の値を用いて学習方法の性能を比較する.
}%

\begin{figure}[t]
  \centering
  \includegraphics[width=0.9\columnwidth]{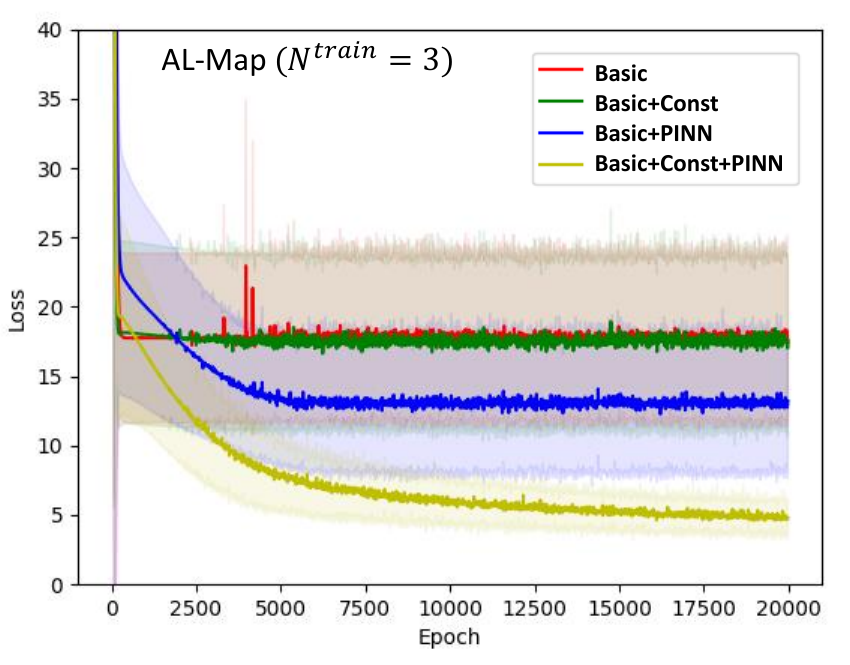}
  \vspace{-1.0ex}
  \caption{The transitions of $L_{eval}$ when training AL-Map with 3 data points in the 2-DOF 4-muscle musculoskeletal simulation.}
  \label{figure:wo-tension-data-03-result}
  \vspace{-2.0ex}
\end{figure}

\begin{figure}[t]
  \centering
  \includegraphics[width=0.9\columnwidth]{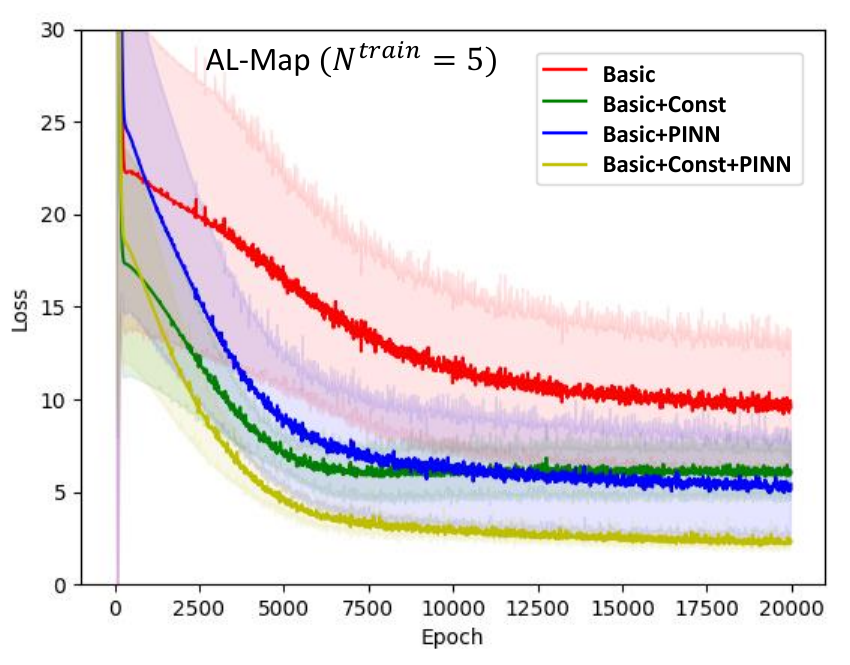}
  \vspace{-1.0ex}
  \caption{The transitions of $L_{eval}$ when training AL-Map with 5 data points in the 2-DOF 4-muscle musculoskeletal simulation.}
  \label{figure:wo-tension-data-05-result}
  \vspace{-3.0ex}
\end{figure}

\begin{figure}[t]
  \centering
  \includegraphics[width=0.9\columnwidth]{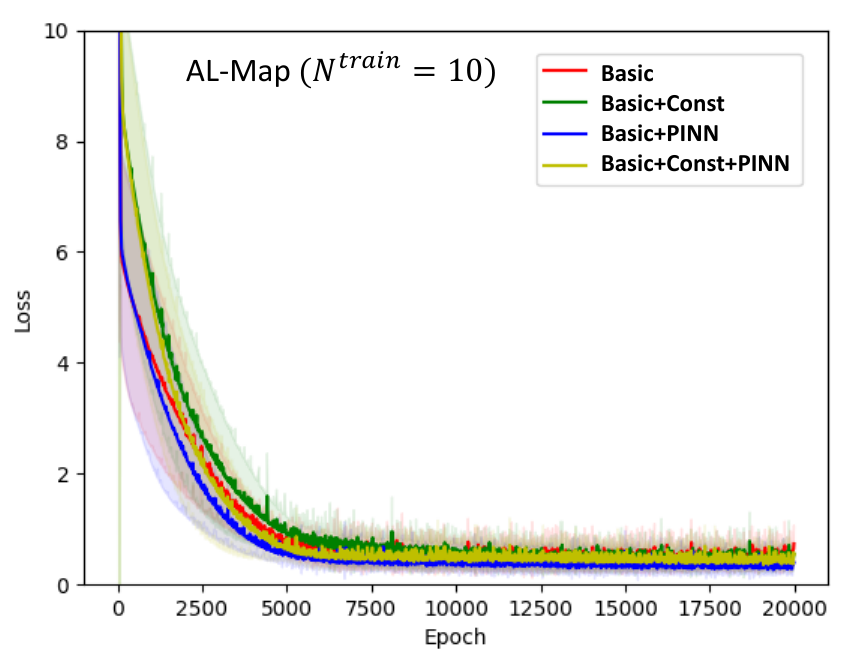}
  \vspace{-1.0ex}
  \caption{The transitions of $L_{eval}$ when training AL-Map with 10 data points in the 2-DOF 4-muscle musculoskeletal simulation.}
  \label{figure:wo-tension-data-10-result}
  \vspace{-3.0ex}
\end{figure}

\begin{figure*}[t]
  \centering
  \includegraphics[width=1.6\columnwidth]{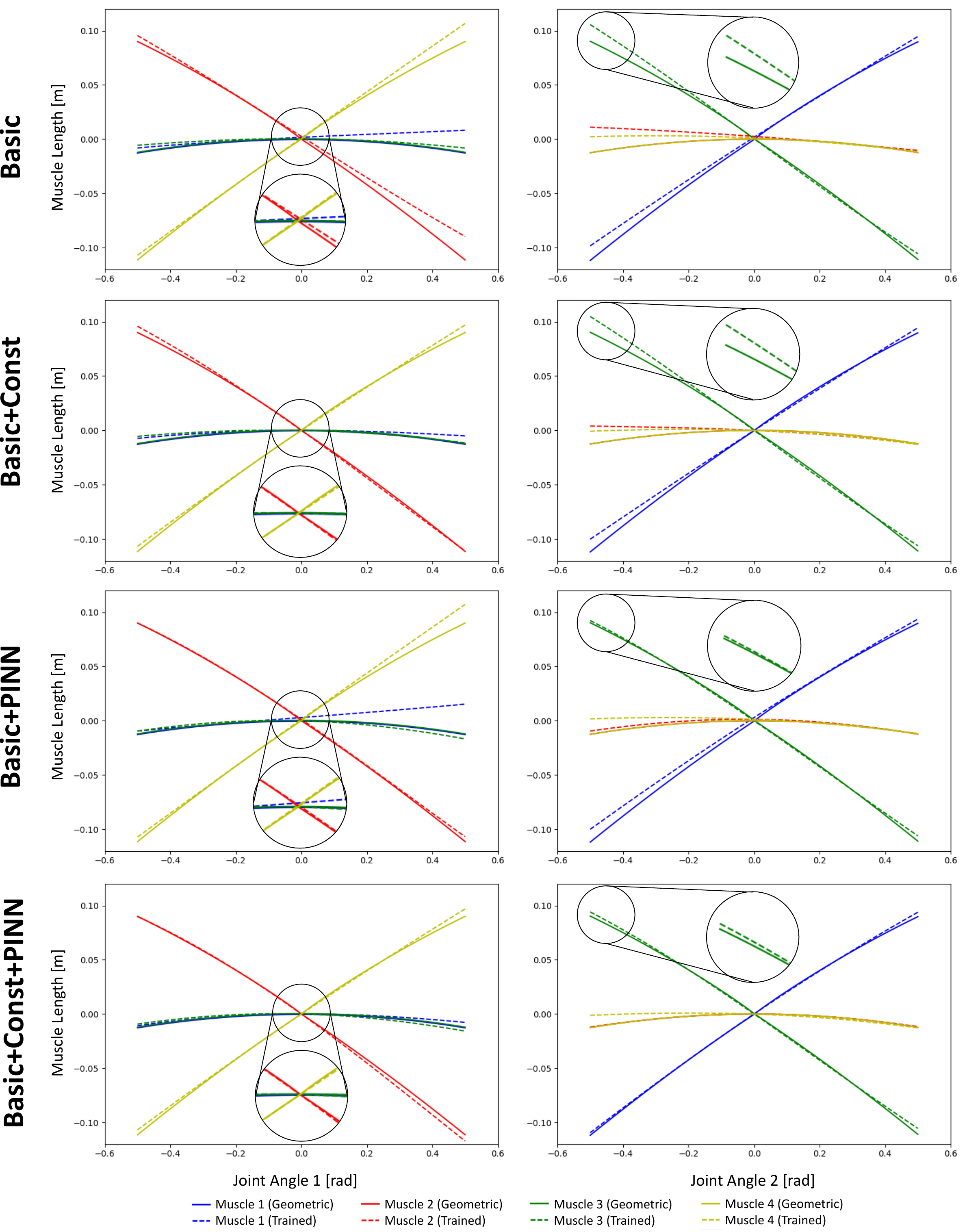}
  \vspace{-1.0ex}
  \caption{The relationship between joint angle and muscle length when training AL-Map with 5 data points in the 2-DOF 4-muscle musculoskeletal simulation.}
  \label{figure:analysis}
  \vspace{-3.0ex}
\end{figure*}

\subsection{Body Schema Learning of AL-Map for 2-DOF 4-muscle Musculoskeletal Simulation} \label{subsec:sim-exp-1}
\switchlanguage%
{%
  First, we describe the results of learning the body schema represented by the AL-Map in a 2-DOF 4-muscle musculoskeletal simulation.
  Here, experiments are conducted by varying $N^{train}=\{3, 5, 10\}$.
  The results for $N^{train}=3$, $N^{train}=5$, and $N^{train}=10$ are shown in \figref{figure:wo-tension-data-03-result}, \figref{figure:wo-tension-data-05-result}, and \figref{figure:wo-tension-data-10-result}, respectively.
  Each experiment was repeated five times with different random seeds, and the average and standard deviation of the transition of $L_{eval}$ are presented.
  This applies to all subsequent graphs as well.

  For $N^{train}=3$, \textbf{Basic} and \textbf{Basic+Const} show similar performance, while both \textbf{Basic+PINN} and \textbf{Basic+Const+PINN} exhibit significant performance improvement.
  For $N^{train}=5$, \textbf{Basic+Const} outperforms \textbf{Basic}, and both \textbf{Basic+PINN} and \textbf{Basic+Const+PINN} further enhance performance.
  For $N^{train}=10$, these differences become negligible.
  As a general trend, increasing $N^{train}$ improves overall performance, but the performance gap between methods decreases.
  Notably, for $N^{train}=5$, both \textbf{Const} and \textbf{PINN} contribute to performance improvement, with \textbf{Basic+Const+PINN} achieving the highest performance.

  \tabref{table:exp-wo-tension} shows the average and standard deviation of $L^{best}_{eval}$ for each learning method.
  For $N^{train}=\{3, 5\}$, \textbf{Basic+Const+PINN} achieves the best performance, reducing $L^{best}_{eval}$ by approximately 60--70\% compared to \textbf{Basic}.
  In contrast, for $N^{train}=10$, performance differences become minimal.
  Additionally, it is evident that \textbf{PINN} has a greater impact on performance than \textbf{Const}.

  \figref{figure:analysis} shows the relationship between joint angle and muscle length for $N^{train}=5$.
  The relationship in the geometric model to be learned is labeled as Geometric, while the relationship derived from the learned body schema is labeled as Trained, for comparison.
  The left figure shows the transition when the first joint is moved, and the right figure shows the transition when the second joint is moved.
  When \textbf{Const} is included in the loss function, the muscle length is learned to become zero near the origin of the joint angle, with the transitions of each muscle intersecting neatly at the origin.
  On the other hand, more errors are observed without \textbf{Const} compared to when it is included.
  When \textbf{PINN} is included in the loss function, performance is improved, particularly at the extremes of the joint angle range.
  Since the joint angle range is randomly varied, data near the extremes are sparse, but utilizing gradient information enables successful extrapolation in these regions.
}%
{%
  まず, 2-DOF 4-muscleの筋骨格シミュレーションにおいて, AL-Mapによって表現された身体図式を学習した際の結果について述べる.
  ここでは, $N^{train}=\{3, 5, 10\}$と変化させながら実験を行う.
  $N^{train}=3$の結果を\figref{figure:wo-tension-data-03-result}, $N^{train}=5$の結果を\figref{figure:wo-tension-data-05-result}, $N^{train}=10$の結果を\figref{figure:wo-tension-data-10-result}に示す.
  ここではランダムシードを変えながら5回の実験を行い, その際の$L_{eval}$の遷移の平均値と標準偏差を示しており, これは今後のグラフも全て同様である.

  $N^{train}=3$のとき, \textbf{Basic}と\textbf{Basic+Const}は同程度の性能を示しているが, \textbf{Basic+PINN}と\textbf{Basic+Const+PINN}は大きく性能が向上している.
  また, $N^{train}=5$のとき, \textbf{Basic+Const}は\textbf{Basic}よりも性能が向上しており, \textbf{Basic+PINN}と\textbf{Basic+Const+PINN}はさらに性能が向上している.
  そして, $N^{train}=10$のとき, これらの違いはほとんど見られなくなっている.
  傾向として, $N^{train}$が増えるに従って性能自体は良くなるが, 手法間の性能差が小さくなっていく.
  特に$N^{train}=5$において顕著であるが, \textbf{Const}にも\textbf{PINN}にも一定の効果があり, これが組み合わさった\textbf{Basic+Const+PINN}が最も性能が高い.

  \tabref{table:exp-wo-tension}には, 各学習方法における$L^{best}_{eval}$の平均値とその標準偏差を示している.
  $N^{train}=\{3, 5\}$においては, \textbf{Basic+Const+PINN}が最も性能が高く, \textbf{Basic}に比べて$L^{best}_{eval}$は60\%から70\%程度下がっている.
  これに対して, $N^{train}=10$のときはほとんど性能差が見られなくなっている.
  また, \textbf{PINN}の方が\textbf{Const}よりも性能への寄与が大きいこともわかる.

  \figref{figure:analysis}には, $N^{train}=5$における関節角度と筋長の関係を示している.
  このとき, 学習すべき幾何モデルにおける関係をGeometric, および学習された身体図式による関係をTrainedとして示し比較している.
  左図を１つ目の関節を動かしたときの遷移, 右図を２つ目の関節を動かしたときの遷移としている.
  \textbf{Const}を損失関数に入れた場合は関節角度が原点付近で筋長が0になるように学習されており, 各筋肉の遷移が原点で綺麗に交差している.
  一方で, \textbf{Const}を入れない場合は, 入れた場合に比べて誤差が乗っていることがわかる.
  \textbf{PINN}を損失に入れた場合は, 特に関節角度範囲の端において, その性能が向上していることがわかる.
  関節角度範囲をランダムに変化させているため, 関節角度の端におけるデータは少ないが, 勾配情報を活用することで外挿に成功していると言える.
}%

\begin{table}[htb]
  \centering
  \caption{Comparison of $L^{best}_{eval}$ and its variance for four training methods of AL-Map in the 2-DOF 4-muscle musculoskeletal simulation.}
  \scalebox{0.95}{
  \begin{tabular}{l|r|r|r}
    Method & $N^{train}=3$ & $N^{train}=5$ & $N^{train}=10$ \\ \hline
    \textbf{Basic}            & 17.75 $\pm$ 6.10 & 9.70 $\pm$ 3.60 & 0.44 $\pm$ 0.21 \\
    \textbf{Basic+Const}      & 17.44 $\pm$ 6.36 & 5.98 $\pm$ 1.37 & 0.42 $\pm$ 0.19 \\
    \textbf{Basic+PINN}       & 12.96 $\pm$ 5.04 & 5.72 $\pm$ 3.41 & \textbf{0.29 $\pm$ 0.12} \\
    \textbf{Basic+Const+PINN} &  \textbf{6.99 $\pm$ 1.86} & \textbf{2.73 $\pm$ 0.38} & 0.47 $\pm$ 0.23 \\
  \end{tabular}
  }
  \label{table:exp-wo-tension}
  \vspace{-3.0ex}
\end{table}

\begin{figure}[t]
  \centering
  \includegraphics[width=0.9\columnwidth]{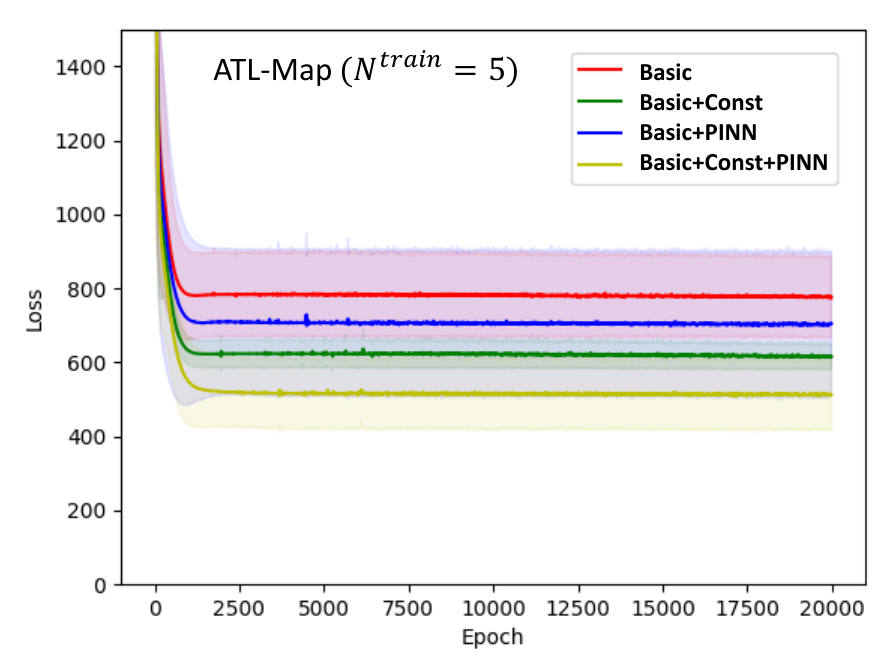}
  \vspace{-2.0ex}
  \caption{The transitions of $L_{eval}$ when training ATL-Map with 5 data points in the 2-DOF 4-muscle musculoskeletal simulation.}
  \label{figure:w-tension-data-05-result}
  \vspace{-3.0ex}
\end{figure}

\begin{figure}[t]
  \centering
  \includegraphics[width=0.9\columnwidth]{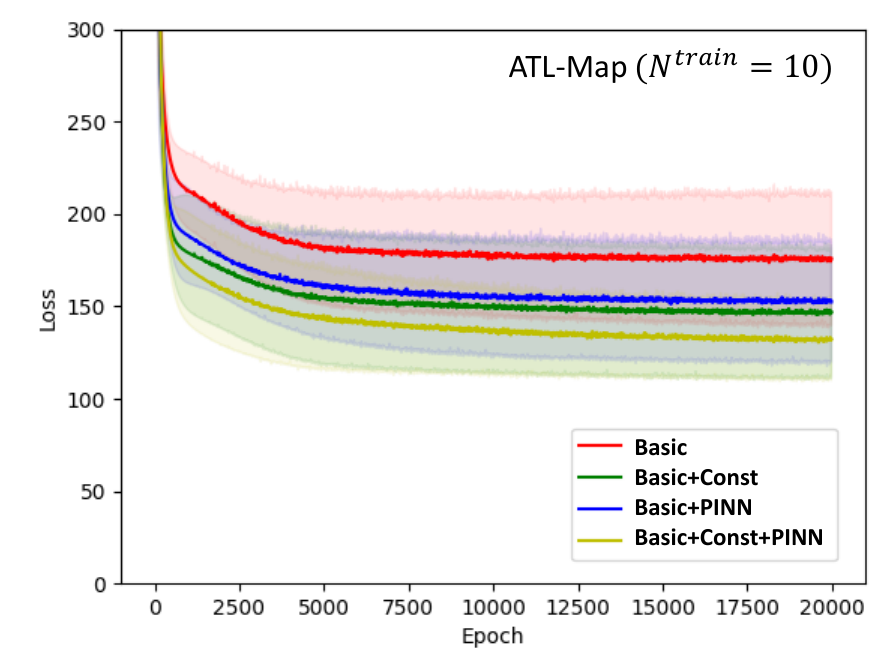}
  \vspace{-2.0ex}
  \caption{The transitions of $L_{eval}$ when training ATL-Map with 10 data points in the 2-DOF 4-muscle musculoskeletal simulation.}
  \label{figure:w-tension-data-10-result}
  \vspace{-3.0ex}
\end{figure}

\begin{figure}[t]
  \centering
  \includegraphics[width=0.9\columnwidth]{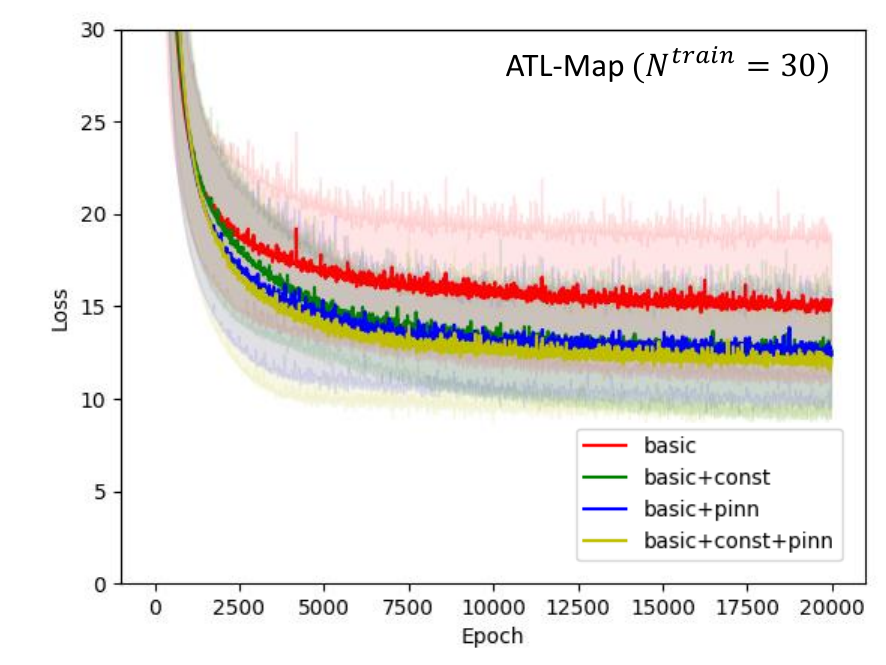}
  \vspace{-2.0ex}
  \caption{The transitions of $L_{eval}$ when training ATL-Map with 30 data points in the 2-DOF 4-muscle musculoskeletal simulation.}
  \label{figure:w-tension-data-30-result}
  \vspace{-3.0ex}
\end{figure}

\subsection{Body Schema Learning of ATL-Map for 2-DOF 4-muscle Musculoskeletal Simulation} \label{subsec:sim-exp-2}
\switchlanguage%
{%
  Next, the results of learning the body schema represented by the ATL-Map in a 2-DOF 4-muscle musculoskeletal simulation are presented.
  Here, experiments are conducted by varying $N^{train} = \{5, 10, 30\}$.
  Since the learning is more challenging than that of AL-Map, the amount of data is set to be greater than in the case of AL-Map.
  The results for $N^{train} = 5$, $N^{train} = 10$, and $N^{train} = 30$ are shown in \figref{figure:w-tension-data-05-result}, \figref{figure:w-tension-data-10-result}, and \figref{figure:w-tension-data-30-result}, respectively.

  For $N^{train} = 5$, the performance is ranked in descending order as follows: \textbf{Basic+Const+PINN}, \textbf{Basic+Const}, \textbf{Basic+PINN}, and \textbf{Basic}.
  For $N^{train} = 10$, a similar trend is observed.
  For $N^{train} = 30$, \textbf{Basic+Const+PINN}, \textbf{Basic+Const}, and \textbf{Basic+PINN} show almost the same performance, with only \textbf{Basic} showing a slight drop in performance.
  Overall, both \textbf{Const} and \textbf{PINN} have a certain positive effect, and the combined method \textbf{Basic+Const+PINN} achieves the highest performance.

  \tabref{table:exp-w-tension} shows the mean and standard deviation of $L^{best}_{eval}$ for each learning method.
  In all cases of $N^{train} = \{5, 10, 30\}$, \textbf{Basic+Const+PINN} achieves the highest performance, with $L^{best}_{eval}$ reduced by approximately 20--35\% compared to \textbf{Basic}.
  The performance gap decreases as $N^{train}$ increases.
  Unlike in the case of the AL-Map, it is also evident that \textbf{Const} contributes more to performance improvement than \textbf{PINN}.
}%
{%
  次に, 2-DOF 4-muscleの筋骨格シミュレーションにおいて, ATL-Mapによって表現された身体図式を学習した際の結果について述べる.
  ここでは, $N^{train}=\{5, 10, 30\}$と変化させながら実験を行う.
  AL-Mapの学習より難しいため, データ数はAL-Mapのときよりも多く設定している.
  $N^{train}=5$の結果を\figref{figure:w-tension-data-05-result}, $N^{train}=10$の結果を\figref{figure:w-tension-data-10-result}, $N^{train}=30$の結果を\figref{figure:w-tension-data-30-result}に示す.

  $N^{train}=5$のとき, \textbf{Basic+Const+PINN}, \textbf{Basic+Const}, \textbf{Basic+PINN}, \textbf{Basic}の順で性能が高い.
  $N^{train}=10$もほとんど同様の傾向を示している.
  $N^{train}=30$のときは, \textbf{Basic+Const+PINN}, \textbf{Basic+Const}, \textbf{Basic+PINN}はほとんど同じ性能を示しており, \textbf{Basic}のみ多少性能が落ちている.
  全体的な傾向として, \textbf{Const}にも\textbf{PINN}にも一定の効果があり, これが組み合わさった\textbf{Basic+Const+PINN}が最も性能が高い.

  \tabref{table:exp-w-tension}には, 各学習方法における$L^{best}_{eval}$の平均値とその標準偏差を示している.
  $N^{train}=\{5, 10, 30\}$の全てにおいて, \textbf{Basic+Const+PINN}が最も性能が高く, \textbf{Basic}に比べて$L^{best}_{eval}$は20\%から35\%程度下がっている.
  なお, その性能差は$N^{train}$が増えるに従って小さくなっている.
  AL-Mapのときと異なり, \textbf{Const}の方が\textbf{PINN}よりも性能への寄与が大きいこともわかる.
}%

\begin{table}[htb]
  \centering
  \caption{Comparison of $L^{best}_{eval}$ and its variance for four training methods of ATL-Map in the 2-DOF 4-muscle musculoskeletal simulation.}
  \scalebox{0.9}{
  \begin{tabular}{l|l|l|l}
    Method & $N^{train}=5$ & $N^{train}=10$ & $N^{train}=30$ \\ \hline
    \textbf{Basic}            & 783.45 $\pm$ 114.46 & 176.25 $\pm$ 34.74 & 14.92 $\pm$ 3.85 \\
    \textbf{Basic+Const}      & 623.38 $\pm$  37.68 & 146.55 $\pm$ 35.31 & 12.37 $\pm$ 3.45 \\
    \textbf{Basic+PINN}       & 706.89 $\pm$ 197.87 & 152.92 $\pm$ 32.71 & 12.49 $\pm$ 2.76 \\
    \textbf{Basic+Const+PINN} & \textbf{515.84 $\pm$  95.48} & \textbf{132.35 $\pm$ 20.99} & \textbf{11.91 $\pm$ 2.89} \\
  \end{tabular}
  }
  \label{table:exp-w-tension}
  \vspace{-3.0ex}
\end{table}

\subsection{Body Schema Learning of ATL-Map for the Actual Musculoskeletal Humanoid} \label{subsec:act-exp}
\begin{figure}[t]
  \centering
  \includegraphics[width=0.9\columnwidth]{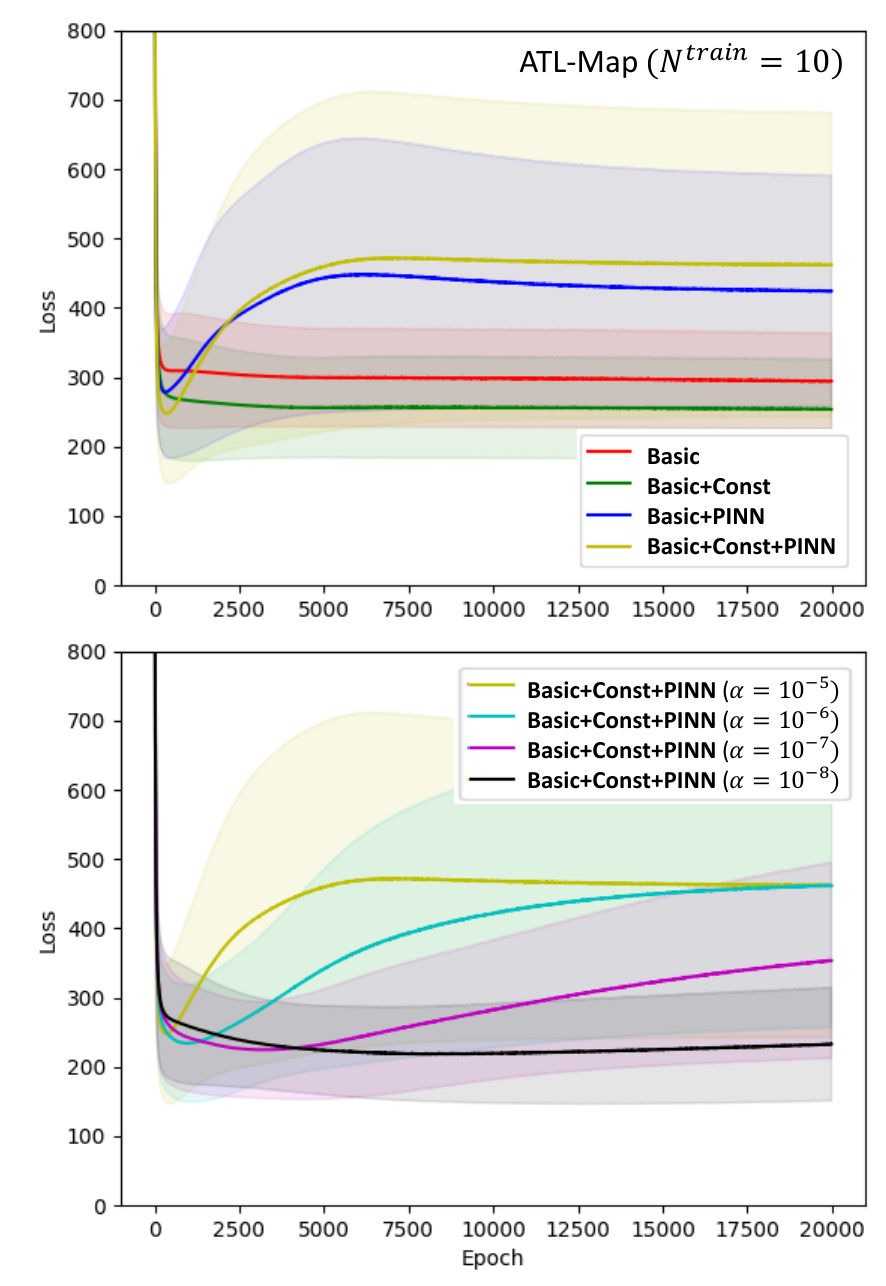}
  \vspace{-2.0ex}
  \caption{The transitions of $L_{eval}$ when training ATL-Map with 10 data points in the left arm of the musculoskeletal humanoid.}
  \label{figure:read-ros-data-10-result}
  \vspace{-3.0ex}
\end{figure}

\begin{figure}[t]
  \centering
  \includegraphics[width=0.9\columnwidth]{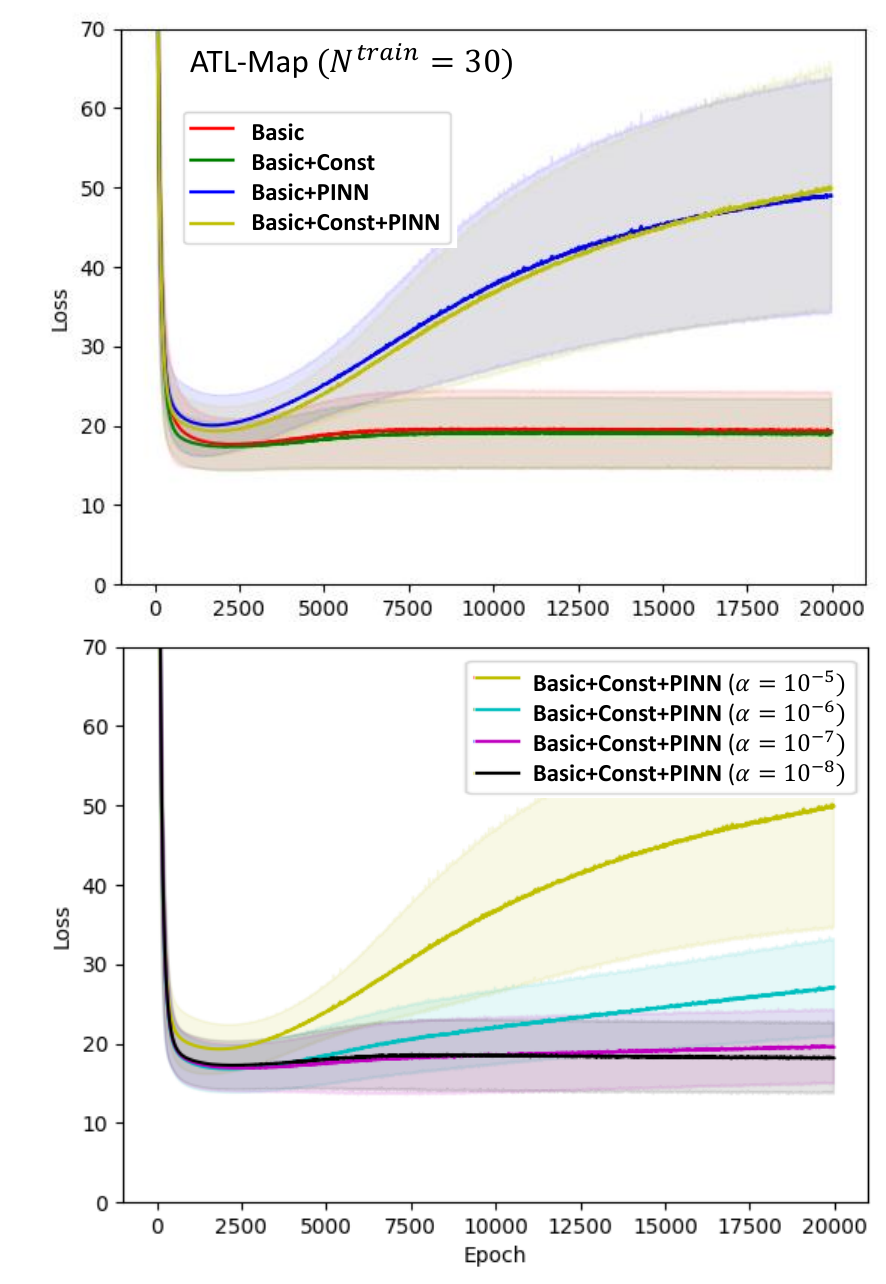}
  \vspace{-2.0ex}
  \caption{The transitions of $L_{eval}$ when training ATL-Map with 30 data points in the left arm of the musculoskeletal humanoid.}
  \label{figure:read-ros-data-30-result}
  \vspace{-3.0ex}
\end{figure}

\switchlanguage%
{%
  Finally, we describe the results of learning the body schema represented by the ATL-Map on the 5-DOF 10-muscle left arm of a musculoskeletal humanoid \cite{kawaharazuka2019musashi}.
  Experiments were conducted by varying $N^{train} = \{10, 30\}$.
  In addition to the results obtained with $\alpha = 1.0 \times 10^{-5}$ -- as used in simulation experiments -- we also show results for \textbf{Basic+Const+PINN} with $\alpha = 1.0 \times 10^{\{-6, -7, -8\}}$.
  For the case of $\alpha = 1.0 \times 10^{-5}$, the results for $N^{train} = 10$ and $30$ are shown in the upper parts of \figref{figure:read-ros-data-10-result} and \figref{figure:read-ros-data-30-result}, respectively.
  For \textbf{Basic+Const+PINN} with $\alpha = 1.0 \times 10^{\{-5, -6, -7, -8\}}$, the results for $N^{train} = 10$ and $30$ are shown in the lower parts of \figref{figure:read-ros-data-10-result} and \figref{figure:read-ros-data-30-result}, respectively.

  First, for $\alpha = 1.0 \times 10^{-5}$, regardless of whether $N^{train} = 10$ or $30$, the performance is highest for \textbf{Basic+Const}, followed by \textbf{Basic}, while adding \textbf{PINN} leads to a degradation in performance over training epochs.
  In contrast, when using smaller values of $\alpha = 10^{\{-6, -7, -8\}}$, this performance degradation is gradually mitigated.

  \tabref{table:exp-read-ros} shows the average and standard deviation of $L^{best}_{eval}$ for each learning method.
  When $N^{train} = 10$, \textbf{Basic+Const+PINN} with $\alpha = 10^{-7}$ achieves the best performance.
  When $N^{train} = 30$, \textbf{Basic+Const+PINN} with $\alpha = 10^{-8}$ performs best.
  Furthermore, a consistent trend is observed in which the performance gap decreases as $N^{train}$ increases.
}%
{%
  最後に, 5-DOF 10-muscleの筋骨格ヒューマノイド \cite{kawaharazuka2019musashi}の左腕において, ATL-Mapによって表現された身体図式を学習した際の結果について述べる.
  ここでは, $N^{train}=\{10, 30\}$と変化させながら実験を行う.
  また, これまでと同様の$\alpha=1.0 \times 10^{-5}$の際の結果だけでなく, \textbf{Basic+Const+PINN}のみ$\alpha=1.0 \times 10^{\{-6, -7, -8\}}$の際の結果も示す.
  $\alpha=1.0 \times 10^{-5}$の場合について, $N^{train}=10$の結果を\figref{figure:read-ros-data-10-result}の上図, $N^{train}=30$の結果を\figref{figure:read-ros-data-30-result}の上図に示す.
  また, $\alpha=1.0 \times 10^{\{-5, -6, -7, -8\}}$として場合の\textbf{Basic+Const+PINN}について, $N^{train}=10$の結果を\figref{figure:read-ros-data-10-result}の下図, $N^{train}=30$の結果を\figref{figure:read-ros-data-30-result}の下図に示す.

  まず$\alpha=1.0 \times 10^{-5}$の場合について, $N^{train}=\{10, 30\}$のどちらにおいても, \textbf{Basic+Const}, \textbf{Basic}の順で性能が高く, \textbf{PINN}を追加することでエポックごとに性能が悪化していることがわかる.
  これに対して, $\alpha=10^{\{-6, -7, -8\}}$として, 物理法則に対する重みを小さくした場合は, $\alpha=1.0 \times 10^{-5}$のときのような性能の悪化が抑えられていく.

  \tabref{table:exp-read-ros}には, 各学習方法における$L^{best}_{eval}$の平均値とその標準偏差を示している.
  $N^{train}=10$のときは\textbf{Basic+Const+PINN} ($\alpha=10^{-7}$)が最も性能が高い.
  $N^{train}=30$のときは\textbf{Basic+Const+PINN} ($\alpha=10^{-8}$)が最も性能が高い.
  また, $N^{train}$が増えるに従って性能差は小さくなるこれまでと同様の傾向が見られる.
}%

\begin{table}[htb]
  \centering
  \caption{Comparison of $L^{best}_{eval}$ and its variance for four training methods of ATL-Map in the left arm of the musculoskeletal humanoid.}
  \scalebox{0.95}{
  \begin{tabular}{l|l|l}
    Method & $N^{train}=10$ & $N^{train}=30$ \\ \hline
    \textbf{Basic}                                & 299.42 $\pm$  70.53 & 19.41 $\pm$  4.84 \\
    \textbf{Basic+Const}                          & 256.57 $\pm$  72.48 & 18.99 $\pm$  4.38 \\
    \textbf{Basic+PINN ($\alpha=10^{-5}$)}        & 423.32 $\pm$ 165.67 & 43.16 $\pm$ 15.87 \\
    \textbf{Basic+Const+PINN ($\alpha=10^{-5}$)}  & 449.73 $\pm$ 217.19 & 44.27 $\pm$ 11.03 \\
    \textbf{Basic+Const+PINN ($\alpha=10^{-6}$)}  & 409.95 $\pm$ 212.65 & 22.65 $\pm$  4.88 \\
    \textbf{Basic+Const+PINN ($\alpha=10^{-7}$)}  & 276.16 $\pm$ 119.48 & \textbf{18.17 $\pm$  4.79} \\
    \textbf{Basic+Const+PINN ($\alpha=10^{-8}$)}  & \textbf{235.31 $\pm$  65.11} & 18.39 $\pm$  4.51 \\
  \end{tabular}
  }
  \label{table:exp-read-ros}
\end{table}

\section{Discussion} \label{sec:discussion}
\switchlanguage%
{%
  The experimental results obtained in this study are summarized here.
  Regarding the simulation experiments, both \textbf{Const} and \textbf{PINN} demonstrated a certain level of effectiveness, and it was found that the combination of these two methods, \textbf{Basic+Const+PINN}, achieved the highest performance.
  Which of \textbf{PINN} or \textbf{Const} has a greater effect depends on the problem setting.
  This difference is more pronounced when $N^{train}$ is small, and as $N^{train}$ increases, the performance gap becomes smaller.
  In contrast, for the actual robot experiments, applying the same weight $\alpha$ for the physical laws as used in simulation led to a degradation in performance when \textbf{PINN} was added.
  However, by reducing $\alpha$, this performance degradation was mitigated, and \textbf{Basic+Const+PINN} once again achieved the highest performance, similar to the simulation results.
  Notably, in simulation, changing the value of $\alpha$ did not result in significant differences in performance.

  Several insights were gained from the experiments.
  By introducing the concept of PINN and utilizing the gradient information within the network, it was possible to ensure a certain level of performance at the boundaries of joint angle limits where the amount of data is scarce.
  When combined with the constraint on returning to the origin imposed by \textbf{Const}, both quantitative and qualitative performance improvements were observed.
  Also, the extent to which physical laws are considered has little effect in simulations, where there is no discrepancy between the data and physical laws.
  However, in the actual robot experiments, factors like friction -- difficult to model as static physical laws -- introduce inconsistencies between the collected data and physical laws.
  Therefore, when physical laws are weighted as heavily as in simulation, the model is overly influenced by incorrect assumptions, leading to a significant drop in performance.
  Reducing the weight for the physical laws helps suppress this degradation while still benefiting somewhat from gradient-based regularization.
  However, even with this adjustment, performance in the actual robot experiments did not reach the level observed in simulation.

  The future outlook for this research is discussed below.
  Although this study demonstrated a certain level of effectiveness, when the influence of friction is significant, such as in the actual robot, modeling that accounts for this effect is crucial.
  In the current static modeling framework, it is difficult to account for frictional effects that depend on the direction of motion.
  Although we attempted a simplified model that considers transmission efficiency, this did not lead to substantial improvements in performance.
  Ideally, a dynamic model that incorporates friction, its resulting dynamics, and hysteresis effects should be constructed.
  However, this would require a greater amount of data, necessitating careful consideration of the balance between modeling complexity and data availability.
  Furthermore, in this study, it was assumed that the joint model was accurate, but in many cases, this assumption may not hold.
  Therefore, future work will aim to develop a formulation capable of handling errors in the joint model as well.

}%
{%
  本研究で得られた実験結果についてまとめる.
  まずシミュレーションについて, \textbf{Const}, \textbf{PINN}のどちらも一定の効果があり, これらが組み合わさった\textbf{Basic+Const+PINN}が最も性能が高いことがわかった.
  \textbf{PINN}か\textbf{Const}のどちらのほうが効果が大きいかは問題設定によって異なる.
  これは$N^{train}$が少ない場合ほど顕著であり, $N^{train}$が増えるに従って性能差は小さくなっていく.
  これに対して実機では, シミュレーションと同様の物理法則に対する重み$\alpha$を使ってしまうと, \textbf{PINN}を追加することで性能が悪化してしまう.
  一方で, $\alpha$を下げることで, この性能悪化を抑え, シミュレーションと同様に\textbf{Basic+Const+PINN}の性能が最も高くなることがわかった.
  なお, シミュレーションでは$\alpha$を変更してもあまり性能差は見られなかった.

  実験からいくつかの示唆が得られた.
  PINNの考え方を導入しネットワークの勾配情報を利用することで, データ数が少ない関節角度限界の端においてある程度の性能を担保することができる.
  \textbf{Const}による原点への制約と合わせて, 定量的にも定性的にも性能が向上することがわかった.
  そして, シミュレーションではデータと物理法則に齟齬がないため, 物理法則をどの程度考慮するかはあまり影響しない.
  一方実機では, 摩擦のような, 静的な状態における物理法則としては考慮の難しい要素が入り, 実機データと物理法則に齟齬が生じてしまう.
  そのため, シミュレーションの際と同程度に物理法則を考慮してしまうと, 間違った物理法則に影響を受け過ぎて, 大きく性能が悪化してしまう.
  これに対して, 物理法則への重みを小さくすることで, 大きな性能悪化を抑えつつ, 勾配情報から多少の性能向上を得ることができた.
  しかし, 物理法則への重みを小さくしたことで, シミュレーションほどの性能は得られなかった.

  今後の展望について述べる.
  本研究は一定の効果があったが, 特に実機のように摩擦の影響が大きい場合には, それを考慮したモデル化が重要となる.
  今回のように静的なモデル化では, 動作の方向に依存した摩擦の影響を考慮することは難しい.
  簡易的に伝達効率を考慮したモデル化を行うことはできるが, それによる性能向上はあまり見られなかった.
  理想的には, 動的なモデルを考え, 摩擦やそれによるダイナミクス, ヒステリシスなどを考慮したモデルを構築することが望ましい.
  一方でこれは, さらなるデータ数を必要とするため, そのバランスを考える必要がある.
  また, 今回は関節のモデルは正確であるという前提をおいているが, これが成り立たない場合も少なくない.
  関節モデルの誤差にも対応可能な定式化を考えていきたい.
}%

\section{CONCLUSION} \label{sec:conclusion}
\switchlanguage%
{%
  In this study, we proposed a new learning method called Physics-Informed Musculoskeletal Body Schema (PIMBS), which incorporates the concept of Physics-Informed Neural Networks (PINNs) for more efficient body schema learning in musculoskeletal humanoids.
  This method leverages not only joint angle, muscle tension, and muscle length data but also the muscle Jacobian obtained through differentiation, as well as the relationship between muscle tension and gravity compensation torque, assuming that the joint structure is correct.
  We demonstrate that this approach enables more efficient body schema learning from a small amount of data with higher performance than conventional methods in both simulation and the actual musculoskeletal humanoid.
  The performance improvement is more pronounced when the amount of data is limited.
  On actual robots, friction can reduce performance; however, we found that adjusting the weight for the physical laws helped mitigate this issue to some extent.
  In this study, we focused on learning the static relationship between joint angle, muscle tension, and muscle length.
  In the future, we aim to extend this to dynamic relationships, incorporating the effects of wire friction, viscosity, and dynamics.
}%
{%
  本研究では, 筋骨格ヒューマノイドのより効率的な身体図式学習に向けた, Physics-Informed Neural Network (PINNs)の考え方を組み込んだ新しい学習方法Physics-Informed Musculoskeletal Body Schema (PIMBS)を提案した.
  関節角度, 筋張力, 筋長のデータだけでなく, これを微分した際に得られる筋長ヤコビアン, 関節構造を正しいと仮定した場合における筋張力と重力補償トルクの関係性を活用した.
  これにより, シミュレーションと実機の筋骨格構造において, 一般的な身体図式学習よりも高い性能をもって少数のデータから身体図式が学習できることを示している.
  その性能差はデータ数が少ない場合ほど顕著である.
  実機では摩擦の影響が性能の低下を招くが, 物理法則への重みを調整することで, ある程度解消出来ることがわかった.
  本研究では静的な関節角度-筋張力-筋長の関係性を学習したが, 今後はこれを動的な関係性へと発展させ, ワイヤの摩擦や粘性,  ダイナミクスを考慮した手法へと昇華させていきたい.
}%

{
  \bibliographystyle{IEEEtran}
  \bibliography{main}
}

\end{document}